\documentclass[twoside,11pt]{article}

\usepackage{jmlr2e}

\usepackage{microtype}
\usepackage{graphicx}
\usepackage{subfigure}
\usepackage{booktabs} 
\usepackage{balance}

\usepackage[table]{xcolor}
\definecolor{Gray}{gray}{0.9}
\usepackage{hhline}

\usepackage{hyperref}

\usepackage{url}

\usepackage[utf8]{inputenc} 
\usepackage[T1]{fontenc}    
\usepackage{scrextend}
\usepackage{booktabs}       
\usepackage{amsfonts}       
\usepackage{amsmath}
\usepackage{amssymb}
\usepackage{mathtools}
\usepackage{color, colortbl}
\usepackage{multirow, multicol}
\usepackage{graphicx}

\jmlrheading{}{}{}{}{}{}

\ShortHeadings{Graph Convolutional Value Decomposition in Multi-Agent Reinforcement Learning}{Naderializadeh, Hung, Soleyman, and Khosla}
\firstpageno{1}

\begin{document}

\title{Graph Convolutional Value Decomposition in \\ Multi-Agent Reinforcement Learning}


\author{\name Navid Naderializadeh \email nnaderializadeh@hrl.com\\
\name Fan H. Hung \email fhhung@hrl.com\\
\name Sean Soleyman \email ssoleyman@hrl.com\\
\name Deepak Khosla \email dkhosla@hrl.com\\
\addr  Information \& Systems Sciences Lab\\
HRL Laboratories, LLC \\
Malibu, CA 90265, USA
}


\maketitle

\begin{abstract}
We propose a novel framework for value function factorization in multi-agent deep reinforcement learning (MARL) using graph neural networks (GNNs). In particular, we consider the team of agents as the set of nodes of a complete directed graph, whose edge weights are governed by an attention mechanism. Building upon this underlying graph, we introduce a mixing GNN module, which is responsible for i) factorizing the team state-action value function into individual per-agent observation-action value functions, and ii) explicit credit assignment to each agent in terms of fractions of the global team reward. Our approach, which we call GraphMIX, follows the centralized training and decentralized execution paradigm, enabling the agents to make their decisions independently once training is completed. We show the superiority of GraphMIX as compared to the state-of-the-art on several scenarios in the StarCraft II multi-agent challenge (SMAC) benchmark. We further demonstrate how GraphMIX can be used in conjunction with a recent hierarchical MARL architecture to both improve the agents' performance and enable fine-tuning them on mismatched test scenarios with higher numbers of agents and/or actions.
\end{abstract}


\section{Introduction}
Multi-agent systems are ubiquitous in numerous application areas, such as autonomous driving~\citep{zhao2019multi, Chu2020Multi-agent}, drone swarms~\citep{zanol2019drone}, communication systems~\citep{9154250}, multi-robot search and rescue~\citep{malaschuk2020intelligent}, and smart grid~\citep{xie2019privacy}. In many of these domains, instructive feedback is not available, as there are no ground-truth solutions or decisions available. These phenomena have given rise to a plethora of literature on multi-agent reinforcement learning algorithms, with a special focus on deep-learning-driven methods over the past few years. 

More recently, algorithms with centralized training and decentralized execution have gained interest, due to their applicability in practical real-world scenarios. In~\citep{lowe2017multi,coma}, policy gradient algorithms are considered, where the actors, which are responsible for taking the actions for each agent, are decentralized, while the critic is assumed to be centralized, trained in conjunction with the actors in an end-to-end manner over the course of training. The authors in~\citep{sunehag2018value, rashid2018qmix} take a different value-based approach to train the agents. Specifically, they introduce a \emph{value factorization} module (linear in the case of VDN~\citep{sunehag2018value} and a state-based non-linear multi-layer perceptron (MLP) in the case of QMIX~\citep{rashid2018qmix}), which is responsible for implicit \emph{credit assignment}; i.e., decomposing the global state-action value function to individual observation-action value functions for different agents.

One of the main drawbacks of the aforementioned algorithms is that they do not explicitly capture the underlying structure of the team of agents in the environment, which can be modeled using a graph topology. There have been some attempts to connect multi-agent deep reinforcement learning (MARL) with graph representation learning methods in the literature. As an example,~\citep{Jiang2020Graph} propose a MARL algorithm based on the graph convolutional network (GCN) architecture~\citep{kipf2017semi}. However, it needs both centralized training and centralized execution (or at the very least, the agents need to communicate with each other multiple times during the inference phase), and therefore, it does not allow for decentralized decision making by the agents.

In this paper, we propose an algorithm for training MARL agents via graph neural networks (GNNs). 
We consider a complete directed weighted graph, where each node represents an agent, and there is a directed edge between any pair of nodes. We use an attention mechanism to dynamically adjust the weights of the edges based on the agents' observation-action histories during an episode. Leveraging such a graph structure, we propose to use a \emph{mixing GNN module} that produces a global state-action value function at its output given the individual agents' observation-action value functions at the input. Similar to~\citep{rashid2018qmix}, a monotonicity constraint is enforced on the GNN, ensuring that the individual agent decisions are consistent with the case where a central entity would be responsible for making the decisions for all the agents.

We also use the mixing GNN as a backbone to derive an effective fraction of global team reward for each of the agents based on their corresponding output node embeddings. We use these reward fractions to minimize per-agent local losses, alongside the global loss using the global state-action value function. The mixing GNN, the attention mechanism, and the agent parameters are all trained centrally in an end-to-end fashion, and after training is completed, each agent can make its decisions in a decentralized manner.

We evaluate our proposed algorithm, which we refer to as GraphMIX, on the StarCraft II multi-agent challenge (SMAC) benchmark~\citep{samvelyan2019starcraft} and demonstrate that it outperforms the state-of-the-art QMIX~\citep{rashid2018qmix} and Weighted QMIX~\citep{rashid2020weighted} algorithms across several hard and super hard scenarios. We also show that GraphMIX can be combined with a recently-proposed hierarchical MARL framework, namely RODE~\citep{wang2021rode}, to provide a further performance improvement over both vanilla GraphMIX and RODE approaches, as well as the capability to fine-tune agents on test scenarios with different numbers of allied/enemy units as compared to the training scenarios. 



\section{Related Work}\label{sec:related_work}

\textbf{Multi-Agent Deep Reinforcement Learning~} A growing focus in the recent deep reinforcement learning literature is on multi-agent cooperation~\citep{ lowe2017multi,coma,sunehag2018value,rashid2018qmix,papoudakis2020comparative}. Methods exist on a spectrum from a single unified or centralized agent to independent and decentralized agents. At one extreme end of this spectrum, a MARL problem might be reduced to a standard deep reinforcement learning problem, with a single centralized network returning a joint action vector for all agents. In this extreme case, issues that arise, such as the exponential increase in the joint action space dimensionality, are general motivations for the development of new algorithms in the MARL literature, as they create difficulties in generalization to different numbers of agents, parameter memory scalability, and training sample efficiency. At the other extreme, independent and decentralized agents face difficulty as coordination becomes more complex. A recent trend that aims to find an effective middle ground is centralized training and decentralized execution (CTDE). Methods in the CTDE paradigm aim to produce decentralized controllers, and enforce implicit coordination with a centralized architecture used only in training. A related trend, which skews more toward a centralized agent, studies the impact of communication between agents during execution~\citep{foerster2016learning, sukhbaatar2016learning}.

\textbf{Value Decomposition Methods for MARL~} Most related to our work is a branch of value-based MARL methods, which decompose a joint state-action value function to allow individual agents to be trained from a single global reward. VDN~\citep{sunehag2018value} initially approximated a joint state-action value function over all agents' actions as the sum of individual observation-action value functions from each agent. QMIX~\citep{rashid2018qmix} observes that the joint state-action value function can more generally be represented by a monotonic function of individual observation-action value functions. Additionally, QMIX allows the joint state-action value function to be informed by global information, which is unavailable to the individual observation-action value functions. The specifics of this factorization continue to be analyzed and improved in several recent works, such as~\citep{son2019qtran, yang2020qatten, rashid2020weighted, wang2021qplex}. In particular, the authors in~\citep{rashid2020weighted} introduce a variation of QMIX, called Weighed QMIX, which expands the representational limits of QMIX by giving a higher weight to more important joint action vectors. Moreover, in~\citep{wang2021qplex}, the authors propose a new duplex dueling value decomposition architecture, called QPLEX, which encompasses a complete class of functions that satisfy the so-called Individual-Global-Max (IGM) principle~\citep{son2019qtran}.

\textbf{Graph Neural Networks~} In order to manage arbitrarily-structured input data that cannot be modeled as regular grids, graph neural networks (GNNs) have gained popularity as a prominent method for incorporating neighborhood information among nodes in graph-based data structures~\citep{scarselli2008graph, kipf2017semi, zhou2018graph, xu2018how}. Similarly to convolutional and recurrent neural networks, GNNs formalize structured treatment of data that would otherwise be concatenated and treated purely as vectors in a high-dimensional space. Due to the flexibility in modeling structured data, GNNs have seen widespread application in areas such as knowledge representation~\citep{park2019estimating}, natural language processing~\citep{ji2019graph}, social network analysis~\citep{fan2019graph}, wireless communications~\citep{9072356}, chemistry~\citep{Hu*2020Strategies}, and physics~\citep{ju2020graph}.

\textbf{GNN-Based MARL Frameworks~} In MARL, GNN-based architectures have recently been used to improve sample efficiency by adding permutation-invariance to multi-agent critics~\citep{liu2020pic} and emphasize observations of neighborhoods in individual agent controllers~\citep{Jiang2020Graph}. Graph structures have also been tied into neural attention modules, especially when using attention mechanisms to compute graph edge weights~\citep{velickovic2018graph, thekumparampil2018attention}. These mechanisms gained popularity in sentence translation tasks for handling associations between structured data components~\citep{vaswani2017attention, devlin2019bert}, and they have seen use in general reinforcement learning as well~\citep{zambaldi2018deep,Baker2020Emergent,iqbal2019actor}.

In this work, we re-visit the analysis of the joint state-action value function to reflect the geometric graphical structure of the multi-agent setting. We also demonstrate how our method can be applied in a recently-proposed hierarchical MARL framework, called RODE~\citep{wang2021rode}, where each agent first selects a role, defined as a cluster of actions having similar effects on the environment dynamics, and it then chooses an action within the action subspace of the selected role. We specifically show how we can factorize both the high-level and the low-level global value functions (for the roles and actions, respectively) via the proposed method to further enhance the performance of the agents.

\section{System Model}\label{sec:sys_model}
We consider a multi-agent environment, where a team of $M$ agents collaborate with each other to solve a cooperative task. In particular, we consider a decentralized partially-observable Markov decision process (Dec-POMDP), represented by a tuple $\langle M, \mathcal{S}, O, \mathcal{Z}, \mathcal{A}, T, R_g, \gamma \rangle$. At each time step $t$, the environment is in global state $s(t)\in\mathcal{S}$, with $\mathcal{S}$ denoting the global state space, and each agent $m\in\{1,\dots,M\}$ receives an observation $o_m(t) = O(s_t,m)\in\mathcal{Z}$, where $\mathcal{Z}$ denotes the per-agent observation space, and $O: \mathcal{S} \times \{1, \dots, M\} \rightarrow \mathcal{Z}$ denotes the per-agent observation function. Upon receiving its observation, each agent $m\in\{1,\dots,M\}$ takes an action $a_m(t) \in\mathcal{A}$, with $\mathcal{A}$ denoting the per-agent action space. These actions will cause the environment to transition to the next state $s(t+1) \sim T\left(s'|s(t), \{a_m(t)\}_{m=1}^M \right)$, with $T: \mathcal{S} \times \mathcal{S} \times \mathcal{A}^M \rightarrow [0, 1]$ denoting the state transition function. This transition is accompanied with a global reward $r_g(t) = R_g\left(s(t), \{a_m(t)\}_{m=1}^M\right)$, where $R_g: \mathcal{S} \times \mathcal{A}^M \rightarrow \mathbb{R}$ denotes the global reward function.

In this setting, the goal of the agents at each time step $t$ is to take a joint set of actions $\{a_m(t)\}_{m=1}^M$ so as to maximize the \emph{discounted cumulative global reward}, defined as $\sum_{t'=t}^\infty \gamma^{t-t'} r_g(t')$. In order to make such decisions, each agent $m\in\{1,\dots,M\}$ is equipped with a policy $\pi_m : \mathcal{A} \times (\mathcal{Z} \times \mathcal{A})^*  \rightarrow [0,1]$ that determines its action given its observation-action history, where $(\mathcal{Z} \times \mathcal{A})^*$ denotes the set of all possible observation-action histories. In particular, the action of agent $m$ at time step $t$ is distributed as $a_m(t) \sim \pi_m\left(a | \tau_m(t)\right)$, where $\tau_m(t) = \left(\{o_m(t')\}_{t'=1}^t, \{a_m(t')\}_{t'=1}^{t-1}\right)$ denotes the set of current and past local observations, as well as past actions of agent $m$ at time step $t$. Letting $\pi$ denote the set of policies of all agents, its induced joint state-action value function is defined as
\begin{align}
Q^\pi\left(s(t), \{a_m(t)\}_{m=1}^M\right) = \mathbb{E}\left[\sum_{t'=t}^\infty \gamma^{t-t'} r_g(t')\right],
\end{align}
where the expectation is taken with respect to the set of future states and actions.

\begin{figure*}[t]
\centering
\includegraphics[trim=.45in 2.7in .75in 1.8in, clip, width=\textwidth]{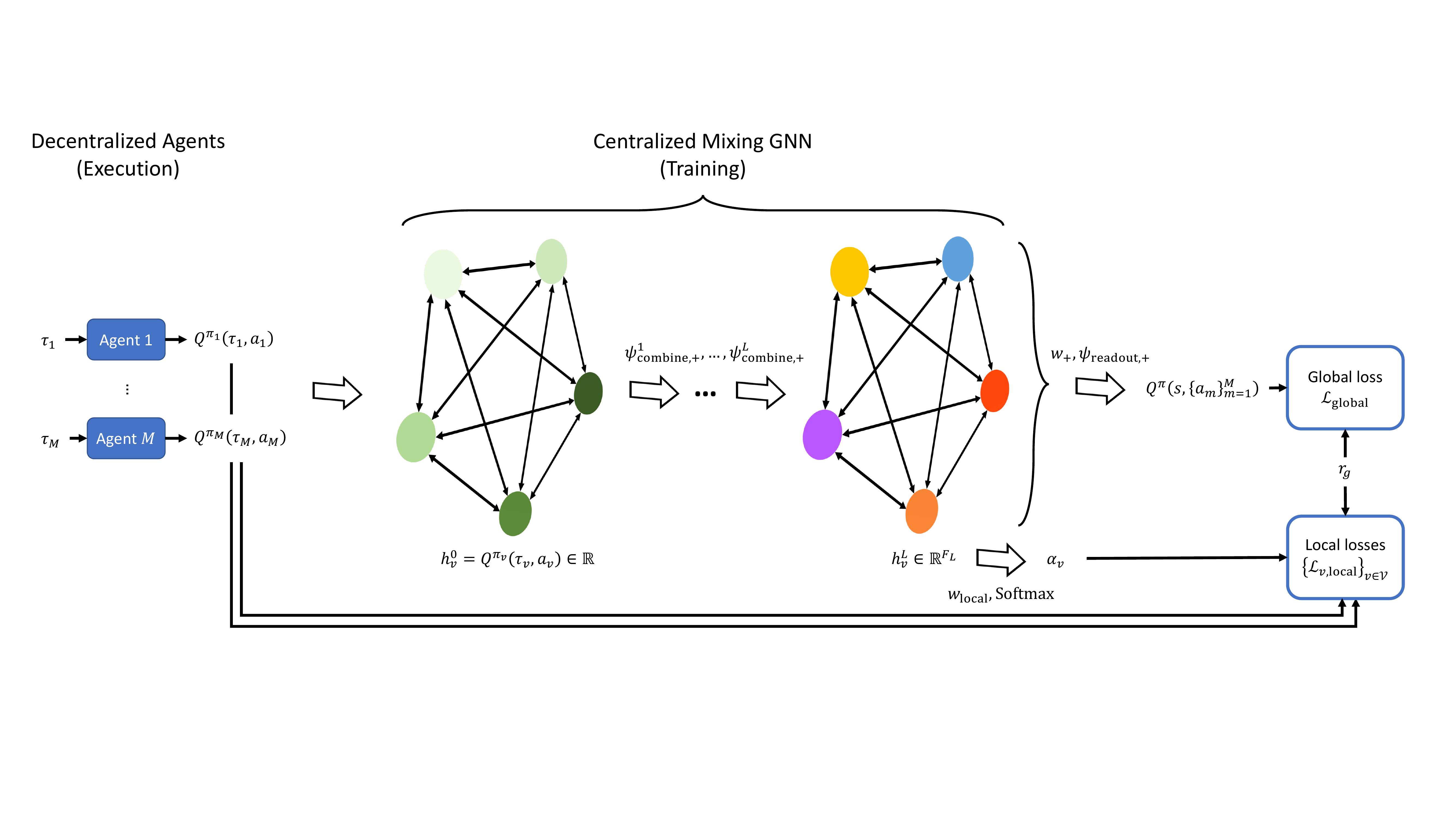}
\caption{The GraphMIX architecture, comprising individual agents that make their decisions in a decentralized manner, alongside a mixing GNN that is used for centralized training of the agents.}
\label{fig:framework}
\end{figure*}

\section{GraphMIX: Graph-Based Value Function Factorization}\label{sec:method}

We assume that each agent $m\in\{1,\dots,M\}$ is equipped with a deep recurrent Q-network (DRQN), coupled with a policy $\pi_m$. At each time step $t$, the agent chooses its action in a decentralized manner based on its current and past local observations, alongside its past actions. In particular, it will choose an action $a_m(t)$ according to the distribution $\pi_m\left(a | \tau_m(t)\right)$. This will lead to its local observation-action value function $Q^{\pi_m}(\tau_m(t), a_m(t))$. Note that such individual value functions are only implicitly defined in the case of a single global reward emitted by the environment, and their implications will become more clear as we describe the value function factorization procedure in this section.

Next, we model the team of agents as a directed graph, denoted by $G=(\mathcal{V}, \mathcal{E})$, where $\mathcal{V}$ denotes the set of $M$ graph nodes, each of which corresponds to an agent. Due to this one-to-one correspondence, hereafter in this paper, we use ``agent'' and ``node'' interchangeably. Moreover, $\mathcal{E} = \mathcal{V}\times \mathcal{V}$ denotes the set of $M^2$ graph edges, implying that the graph is a complete directed graph. For two agents $u,v\in\mathcal{V}$, we let $w_{uv}$ denote the weight of the edge from node $v$ to node $u$. These edge weights can possibly be varying over time, and we will later describe how they are determined.

To train the agents, we borrow the notion of monotonic value function factorization from~\citep{rashid2018qmix}, where the idea is to decompose the global state-action value function $Q^\pi\left(s(t), \{a_m(t)\}_{m=1}^M\right)$ into a set of local observation-action value functions $\{Q^{\pi_m}(\tau_m(t), a_m(t))\}_{m=1}^M$ such that an increase in the local observation-action value function of each agent leads to a corresponding increase in the global state-action value function. To be precise, the decomposition can be written as
\begin{align*}
&Q^\pi\left(s(t), \{a_m(t)\}_{m=1}^M\right) = \Psi_{\mathsf{mix}}\Big( Q^{\pi_1}(\tau_1(t), a_1(t)), \dots, Q^{\pi_M}(\tau_M(t), a_M(t)) \Big),
\end{align*}
where the \emph{mixing function} $\Psi_{\mathsf{mix}}$ satisfies
\begin{align}\label{eq:monotonic}
\frac{\partial \Psi_{\mathsf{mix}}(x_1, \dots, x_M)}{\partial x_m} \geq 0, \forall m\in\{1,\dots,M\}.
\end{align}
The monotonicity condition in \eqref{eq:monotonic} is a sufficient condition for the Individual-Global-Max (IGM) principle, defined by~\citep{son2019qtran}, which ensures that if each agent takes the action that maximizes its local observation-action value function, it would also be the best action for the entire team.

In this work, we propose to use a graph-based approach for combining the local per-agent observation-action value functions into the global state-action value function. In particular, we leverage the aforementioned graph of the agents $G$ to define a \emph{mixing GNN} architecture, as shown in Figure~\ref{fig:framework}. In this GNN, each node $v\in\mathcal{V}$ starts with a scalar feature, which is the corresponding agent's local observation-action value function, i.e.,
\begin{align}
h^0_{v} = Q^{\pi_v}(\tau_v, a_v),
\end{align}
where we have dropped the time dependence for brevity. The features are then passed through $L$ hidden layer(s). At the $l$\textsuperscript{th} layer, $l\in\{1,\dots,L\}$, the features of each node $v\in\mathcal{V}$ are updated as
\begin{align}\label{eq:combining_GNN}
h^l_{v} = \psi_{\mathsf{combine},+}^l \left(h^{l-1}_{v}, \{h^{l-1}_{u}, w_{uv}\}_{u\in\mathcal{V}\setminus\{v\}} \right),
\end{align}
where $\psi_{\mathsf{combine},+}^l(\cdot)$ denotes a monotonically-increasing (and potentially non-linear) parametrized combining function. This implies that each node uses its own features and the other agents' features, alongside its outgoing edge weights to map its input feature (vector) of dimension $F_{l-1}$ to an output (vector) of dimension $F_l$, with $F_0=1$.

At the output of the $L$\textsuperscript{th} layer, each node $v\in\mathcal{V}$ will end up with a feature vector, i.e., \emph{node embedding}, $h^L_{v} \in \mathbb{R}^{F_{L}}$. We then define the global state-action value function as
\begin{align}\label{eq:readout_GNN}
Q^\pi\left(s, \{a_m\}_{m=1}^M\right) = w_+^T ~ \psi_{\mathsf{readout}}(\{h^L_{v}\}_{v\in\mathcal{V}}),
\end{align}
where $\psi_{\mathsf{readout}}: \overbrace{\mathbb{R}^{F_{L}} \times \dots \times \mathbb{R}^{F_{L}}}^M \rightarrow \mathbb{R}^{F_{L}}$ is a graph readout operation (such as average/max pooling), and $w_+\in\mathbb{R}_+^{F_{L}}$ is a non-negative parameter vector that maps the graph embedding $\psi_{\mathsf{readout}}(\{h^L_{v}\}_{v\in\mathcal{V}})$ into the global state-action value function. Note that the monotonicity of $\psi_{\mathsf{combine},+}^l(\cdot)$ in \eqref{eq:combining_GNN} and the non-negativity of $w_+$ in \eqref{eq:readout_GNN} guarantee that the mixing monotonicity condition in \eqref{eq:monotonic} is satisfied.

In addition, we introduce another weight vector $w_{\mathsf{local}}\in\mathbb{R}^{F_{L}}$ that maps each node embedding to an \emph{effective reward fraction} for the corresponding agent, defined as
\begin{align}
\alpha_v &= \mathsf{Softmax}_{\mathcal{V}} \Big(w_{\mathsf{local}}^T ~ h_v^L \Big)  = \frac{\exp\big(w_{\mathsf{local}}^T ~ h_v^L\big)}{\sum_{u\in\mathcal{V}} \exp\big(w_{\mathsf{local}}^T ~ h_u^L\big)}.\label{eq:effective_rewards}
\end{align}
We interpret these values as the effective fraction of the global reward that each agent receives at each time step. The significance of $w_{\mathsf{local}}$ lies in the fact that its parameters do not need to be non-negative, hence improving the expressive power of the mixing GNN beyond monotonic functions.

Our proposed architecture, which we refer to as GraphMIX, is trained by minimizing the aggregate loss
\begin{align}\label{eq:aggregate_loss}
\mathcal{L}  = \mathcal{L}_{\mathsf{global}} + \lambda_{\mathsf{local}} \sum_{v\in\mathcal{V}} \mathcal{L}_{v, \mathsf{local}},
\end{align}
with the global loss defined as
\begin{align}
\mathcal{L}_{\mathsf{global}} &= \sum_{i\in\mathcal{B}} \Bigg[\bigg( \left(r_g + \gamma \max_{a'_1,\dots,a'_M} Q^\pi\left(s', \{a'_m\}_{m=1}^M\right)\right) -Q^\pi\left(s, \{a_m\}_{m=1}^M\right) \bigg)^2\Bigg]_i,\label{eq:global_loss}
\end{align}
where $\mathcal{B}$ denotes a batch of transitions that are sampled from the experience buffer at each round of training, and $s'$ and $\{a'_m\}_{m=1}^M$ respectively correspond to the environment state and agents' actions in the following time step. Moreover, for each node $v\in\mathcal{V}$, the local loss is defined as
\begin{align}
\mathcal{L}_{v, \mathsf{local}} &= \sum_{i\in\mathcal{B}} \Bigg[\bigg( \left(\alpha_v r_g + \gamma \max_{a'_v} Q^{\pi_v}(\tau'_v, a'_v)\right) - Q^{\pi_v}(\tau_v, a_v) \bigg)^2\Bigg]_i,\label{eq:local_loss}
\end{align}
where $\tau'_v$ denotes the observation-action history of the agent corresponding to node $v$ at the following time step. Note how minimizing the local losses in~\eqref{eq:local_loss} creates a shortcut for backpropagating gradients to the individual agent networks, compared to the alternative path through the mixing GNN by minimizing the global loss in~\eqref{eq:global_loss}. Moreover, such local updates of the agent networks maintain the consistency of the global policy and the decentralized agent policies. The balance between the two loss types is attained through a hyperparameter $\lambda_{\mathsf{local}}$ in \eqref{eq:aggregate_loss}, which needs to be tuned to optimize the agents' performance in any given environment.

Similar to~\citep{rashid2018qmix}, we use a hypernetwork architecture to determine the parameters of the mixing GNN during training. In particular, each of the mixing GNN parameter matrices/vectors is derived by reshaping and taking the absolute values of the outputs of a multi-layer perceptron, which takes the global state as the input, hence also satisfying the monotonicity constraint for the mixing GNN.

\subsection{Attention-Based Edge Weights}\label{sec:attn}
Inspired by the deep implicit coordination graph (DICG) architecture proposed by~\citep{li2020deep}, we use an attention mechanism to define the edge weights of the graph $G$, i.e., $\{w_{uv}\}_{(u,v)\in\mathcal{E}}$. To capture the entire observation-action history of the agents, we leverage the hidden states of the agents' DRQNs for deriving the attention weights. In particular, we assume that for each agent $m\in\{1,\dots,M\}$, its policy $\pi_m$ internally uses a function $\omega_m: (\mathcal{Z} \times \mathcal{A})^*  \rightarrow \mathbb{R}^D$ to map the agent's observation-action history $\tau_m$ to a hidden state $c_m = \omega_m(\tau_m)$, where we have dropped the time dependence for brevity. We first use a shared encoder mechanism $\phi: \mathbb{R}^D \rightarrow \mathbb{R}^{D'}$ to encode each hidden state $c_m$ to a $D'$-dimensional embedding $\phi(c_m)$. Then, for each pair of agents $m', m\in\{1,\dots,M\}$, the weight of the edge from agent $m$ to agent $m'$ is defined as
\begin{align}
w_{m'm} &= \mathsf{Softmax}_{m'} \left(\tfrac{\big( W_Q\phi(c_{m'})\big)^T \big( W_K\phi(c_m)\big)}{\phantom{\big(}\sqrt{D'}\phantom{\big)}} \right)\label{eq:attn}
\end{align}
where $W_Q, W_K\in\mathbb{R}^{D' \times D'}$ denote the query and key matrices, respectively, whose parameters are trained in an end-to-end fashion alongside those of the agents, the shared encoder, and the mixing GNN. The softmax operator in~\eqref{eq:attn} ensures that the weights of the \emph{outgoing} edges from each node to the other nodes sum up to unity.


\section{Experimental Evaluation}
\label{sec:result}

\subsection{Setup}\label{sec:experimental_setup}

We evaluate GraphMIX on the StarCraft II multi-agent challenge (SMAC) environment \citep{samvelyan2019starcraft}, which provides a set of different micromanagement challenges for benchmarking distributed multi-agent reinforcement learning methods\footnote{We run all our experiments on version \texttt{SC2.4.10} of StarCraft II, which might explain the discrepancy between our results and the results reported in the literature. The implementation code for GraphMIX is available at \url{https://github.com/navid-naderi/GraphMIX}.}. We specifically consider the set of nine maps, which have been classified as either hard or super-hard by~\citep{samvelyan2019starcraft}. In each map, the \emph{allied} team of agents are controlled by the MARL policy, while the enemy units are controlled by the game’s built-in AI.

We train the agents for a total of $2\times10^6$ time steps, except for four super hard maps (\texttt{6h\_vs\_8z}, \texttt{corridor}, \texttt{3s5z\_vs\_3s6z}, and \texttt{27m\_vs\_30m}), where the training time is extended to $5\times10^{6}$ time steps~\citep{wang2021rode}. Moreover, every $2\times10^4$ time steps, training is paused and the agents are evaluated on a set of 32 test episodes. On four maps (\texttt{corridor}, \texttt{2c\_vs\_64zg}, \texttt{bane\_vs\_bane}, and \texttt{3s\_vs\_5z}), we set the local loss coefficient in~\eqref{eq:aggregate_loss} to $\lambda_{\mathsf{local}}=1$, while for the rest of the maps, it is set to zero.

For the mixing GNN, we use the graph isomorphism network (GIN) architecture~\citep{xu2018how}, where the combining operation in~\eqref{eq:combining_GNN} is given by
\begin{align}~\label{eq:combining_simplified}
h_v^l = \mathsf{MLP}_+^l\left(\sum_{u\in \mathcal{V}} w_{uv}h_u^{l-1}\right), \forall v\in\mathcal{V}, \forall l\in\{1,\dots,L\},
\end{align}
where $\mathsf{MLP}_+^l:F_{l-1}\rightarrow F_l$ denotes a multi-layer perceptron (MLP) with non-negative weights. Further details on the experiments can be found in Appendix~\ref{appx:implementation}.

\subsection{Baseline Methods}\label{sec:baselines}
We compare the performance of GraphMIX against three state-of-the baseline methods of QMIX~\citep{rashid2018qmix}, (Optimistic-)Weighted QMIX~\citep{rashid2020weighted} and RODE~\citep{wang2021rode}. As opposed to QMIX and Weighted QMIX, which are specifically focused on value factorization, RODE presents a hierarchical framework for assigning roles to agents over the course of an episode, where each role is defined as a subset of actions that have a similar impact on the environment. In its architecture, RODE uses two mixing networks for roles and actions, both of which are trained using QMIX. As an extension of GraphMIX and RODE, we also present the results for a \emph{combined RODE + GraphMIX} approach, where GraphMIX is used for credit assignment for both agents' roles and actions. The existence of two mixing GNNs implies that the loss function in~\eqref{eq:aggregate_loss} will have additional terms for the local and global role mixing GNN, as well as an additional role local loss hyperparameter $\lambda_{\mathsf{role}, \mathsf{local}}$. The implementation details of the combined RODE + GraphMIX method, as well as the three baselines, can be found in Appendix~\ref{appx:implementation}.

\begin{figure*}[t]
\centering
\includegraphics[width=\textwidth]{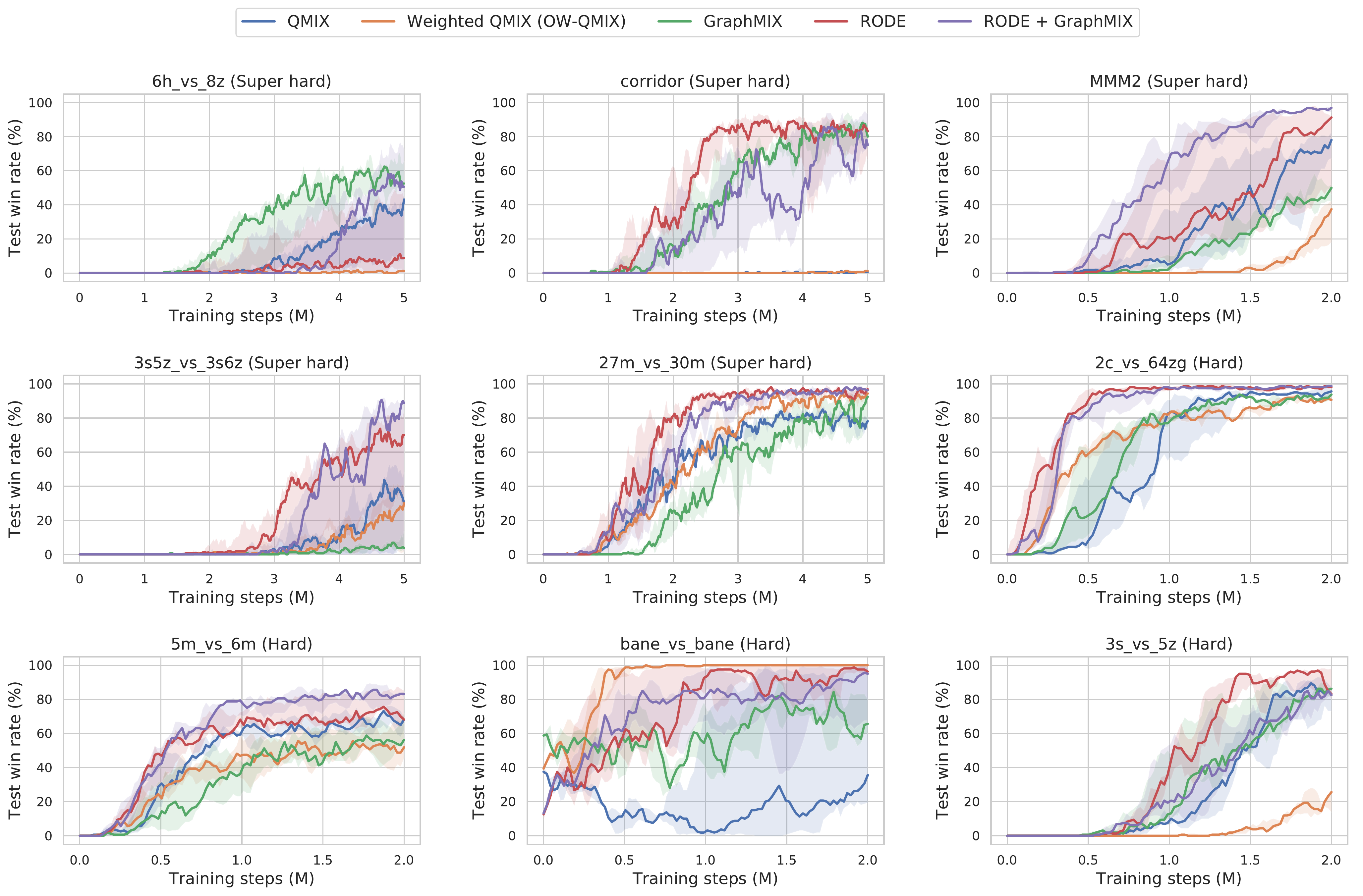}
\caption{Comparison of the test win rates achieved by GraphMIX, combined RODE + GraphMIX, and the three baseline methods on all super hard and hard SMAC maps.}
\label{fig:results_allmaps}
\end{figure*}

\subsection{Results}
Figure~\ref{fig:results_allmaps} shows the test win rates achieved by GraphMIX, combined RODE + GraphMIX, and the three baseline approaches on all the five super hard and four hard SMAC scenarios. The solid curves show the median across five training runs with different random seeds, and the shaded areas represent the 25-75 percentiles. As the figure demonstrates, GraphMIX is able to considerably outperform QMIX and Weighted QMIX on four maps (\texttt{6h\_vs\_8z}, \texttt{corridor}, \texttt{27m\_vs\_30m}, and \texttt{bane\_vs\_bane}) and five maps (\texttt{6h\_vs\_8z}, \texttt{corridor}, \texttt{MMM2}, \texttt{5m\_vs\_6m}, and \texttt{3s\_vs\_5z}), respectively, and it also outperforms RODE on the super hard \texttt{6h\_vs\_8z} map. The combined RODE + GraphMIX approach further boosts the performance, achieving state-of-the-art results on three maps (\texttt{6h\_vs\_8z}, \texttt{MMM2}, and \texttt{3s5z\_vs\_3s6z}).

\begin{figure*}[t]
\centering
\includegraphics[width=\textwidth]{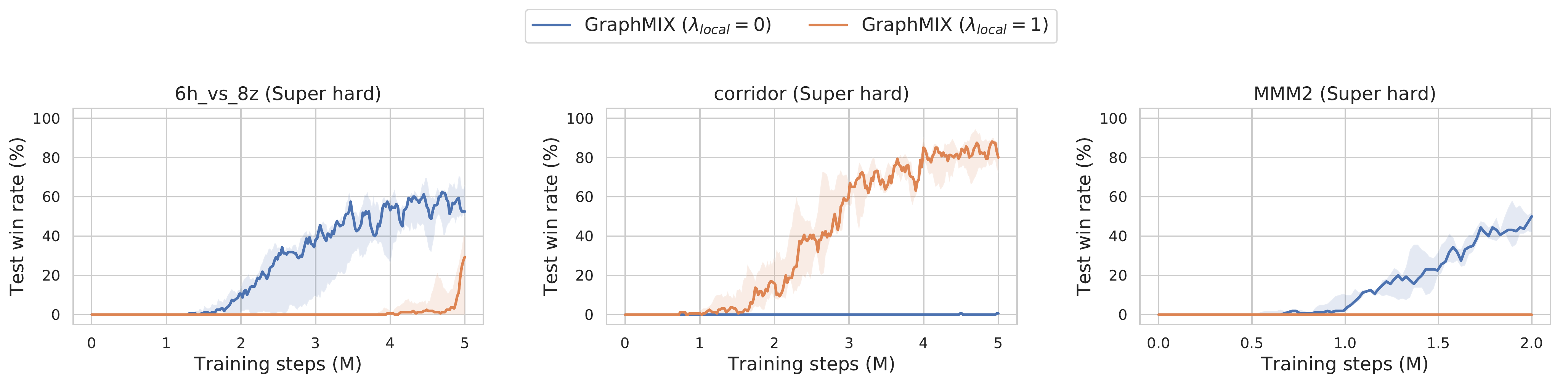}
\caption{Comparison of the performance of GraphMIX on three super hard maps with and without the local loss terms in~\eqref{eq:aggregate_loss}.}
\label{fig:ablation_localloss_threemaps}
\end{figure*}
\begin{figure*}[t]
\centering
\includegraphics[width=\textwidth]{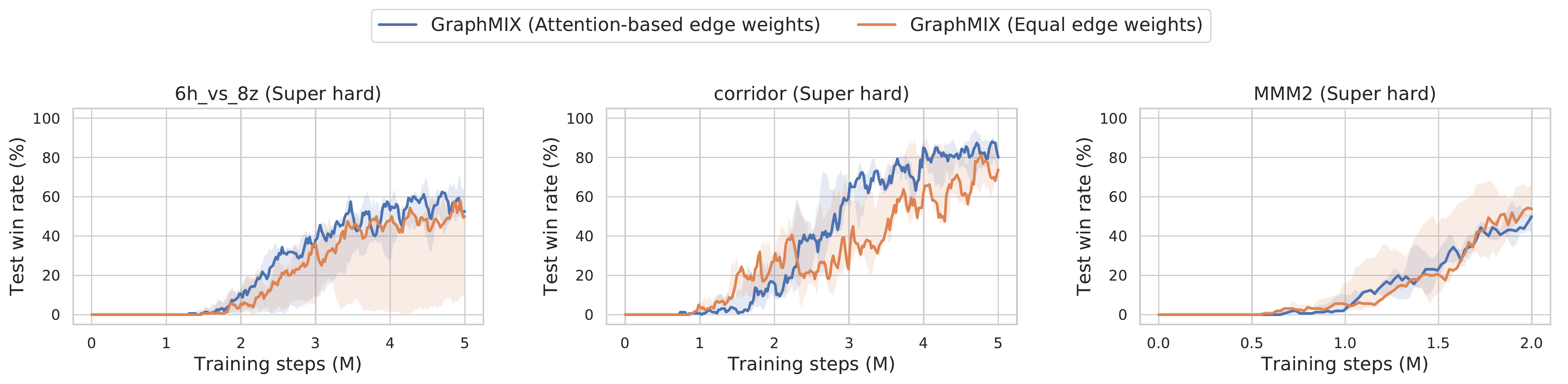}
\caption{Ablation results on the impact of the attention mechanism on the performance of GraphMIX across three super hard maps.}
\label{fig:ablation_attnedgeweights_threemaps}
\end{figure*}

As we mentioned in Section~\ref{sec:experimental_setup}, the local loss coefficient $\lambda_{\mathsf{local}}$ in~\eqref{eq:aggregate_loss} was tuned to either zero or one for each of the nine maps. Figure~\ref{fig:ablation_localloss_threemaps} shows the impact of including (i.e., $\lambda_{\mathsf{local}}=1$) or excluding (i.e., $\lambda_{\mathsf{local}}=0$) the local loss term in~\eqref{eq:aggregate_loss} on the performance of GraphMIX in three super hard maps, namely \texttt{6h\_vs\_8z}, \texttt{corridor}, and \texttt{MMM2}. As the figure demonstrates, the value of $\lambda_{\mathsf{local}}$ has a significant impact on the performance of the proposed method, and therefore, it should be tuned based on the specific dynamics of any given environment to optimize the performance of GraphMIX. The complete performance comparison of GraphMIX with and without the local loss terms on all the nine maps can be found in Appendix~\ref{appx:loss_ablation}.

\begin{figure*}[t]
\centering
\includegraphics[width=\textwidth]{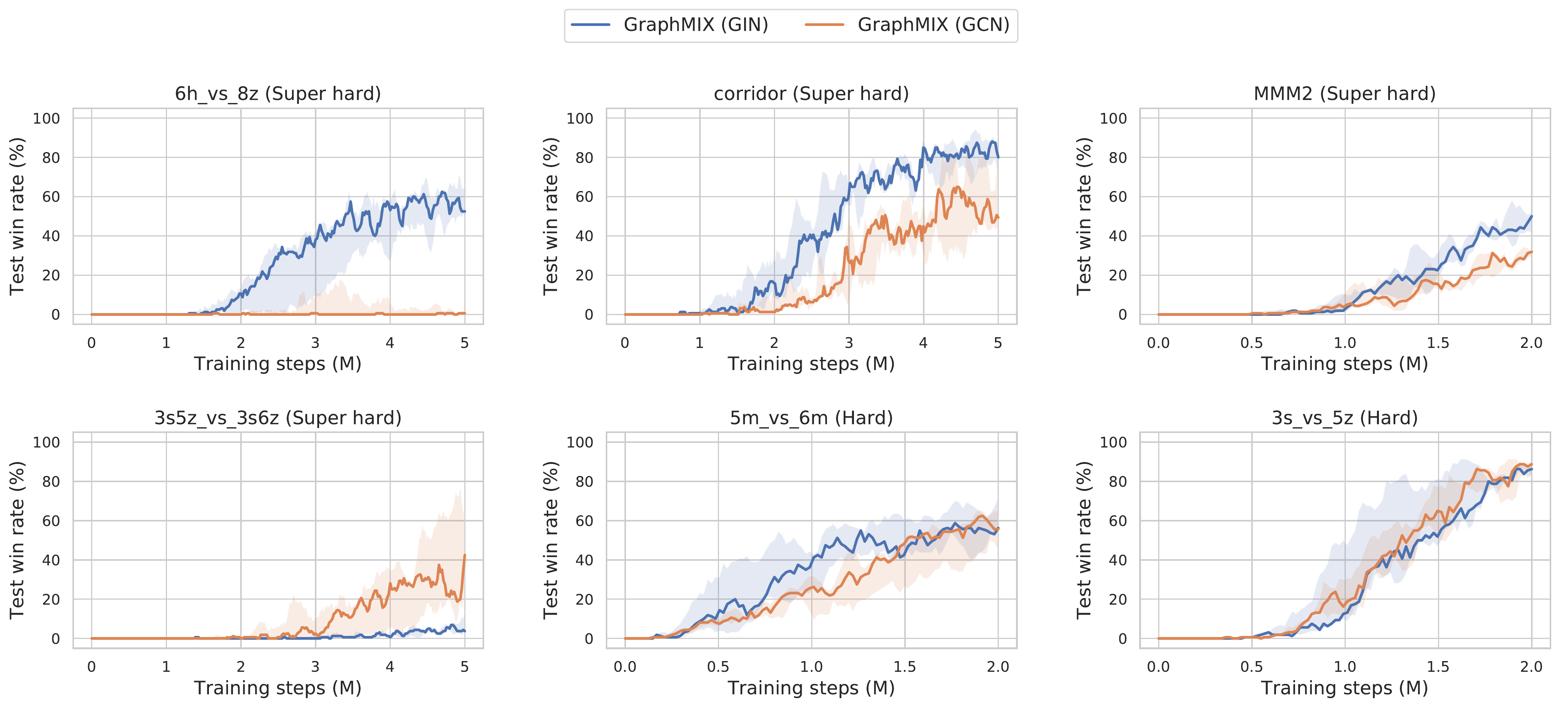}
\caption{Comparison of the performance of GraphMIX under GIN~\citep{xu2018how} and GCN~\citep{kipf2017semi} architectures for the mixing GNN on six hard and super hard maps.}
\label{fig:ablation_GNNarchitecture_fivemaps}
\end{figure*}

Figure~\ref{fig:ablation_attnedgeweights_threemaps} shows the impact of the attention mechanism for determining the edge weights as in~\eqref{eq:attn} on the performance of GraphMIX. In particular, we show the performance of a special case, where all the mixing graph edges have equal weights of $\frac{1}{M}$ (or more precisely, $\frac{1}{\text{\# alive agents}}$). As the figure demonstrates, even though the attention mechanism does help GraphMIX attain a better performance or faster convergence, its contribution is not as significant as the contribution of the local loss term, and it even slightly hurts the performance on \texttt{MMM2}. We hypothesize that this is due to the heterogeneity of the \texttt{MMM2} map, as opposed to the other two maps, which are homogeneous, and more sophisticated attention architectures (in which the attention parameters are shared among agents of similar type, but are different between agents of dissimilar types) can potentially further boost the performance. Note that for the case of equal edge weights, the GIN update in~\eqref{eq:combining_simplified} is simplified to
\begin{align}~\label{eq:combining_simplified_noAttn}
\hspace{-.11in}h_v^l = \mathsf{MLP}_+^l\left(\sum_{u\in \mathcal{V}} h_u^{l-1}\right), \forall v\in\mathcal{V}, \forall l\in\{1,\dots,L\},
\end{align}
where the constant edge weights are absorbed into the MLP. 
Quite interestingly, this can be viewed as an extension of the value decomposition network (VDN) proposed by~\citep{sunehag2018value}, where the only difference is the existence of the non-linear MLP in~\eqref{eq:combining_simplified_noAttn}. Therefore, GraphMIX can also be viewed as a generalization of the size-invariant and permutation-invariant VDN value factorization approach, which can adapt to different environment dynamics thanks to the underlying attention mechanism.

Finally, Figure~\ref{fig:ablation_GNNarchitecture_fivemaps} shows the impact of the mixing GNN architecture on the performance of GraphMIX. In particular, we compare the performance of the GIN architecture with that of the graph convolutional network (GCN) architecture~\citep{kipf2017semi}. As the figure shows, the GIN-based GraphMIX approach is on par with or better than the GCN-based GraphMIX approach on most of the maps, except for \texttt{3s5z\_vs\_3s6z}. This shows that while GIN demonstrates promising performance across all the different maps, the GNN architecture provides another degree of freedom that can be tuned using different GNN combining methods to further boost the performance of GraphMIX.

\subsection{Fine-Tuning on Mismatched Scenarios}
\begin{figure}[t]
\centering
\includegraphics[width=.7\linewidth]{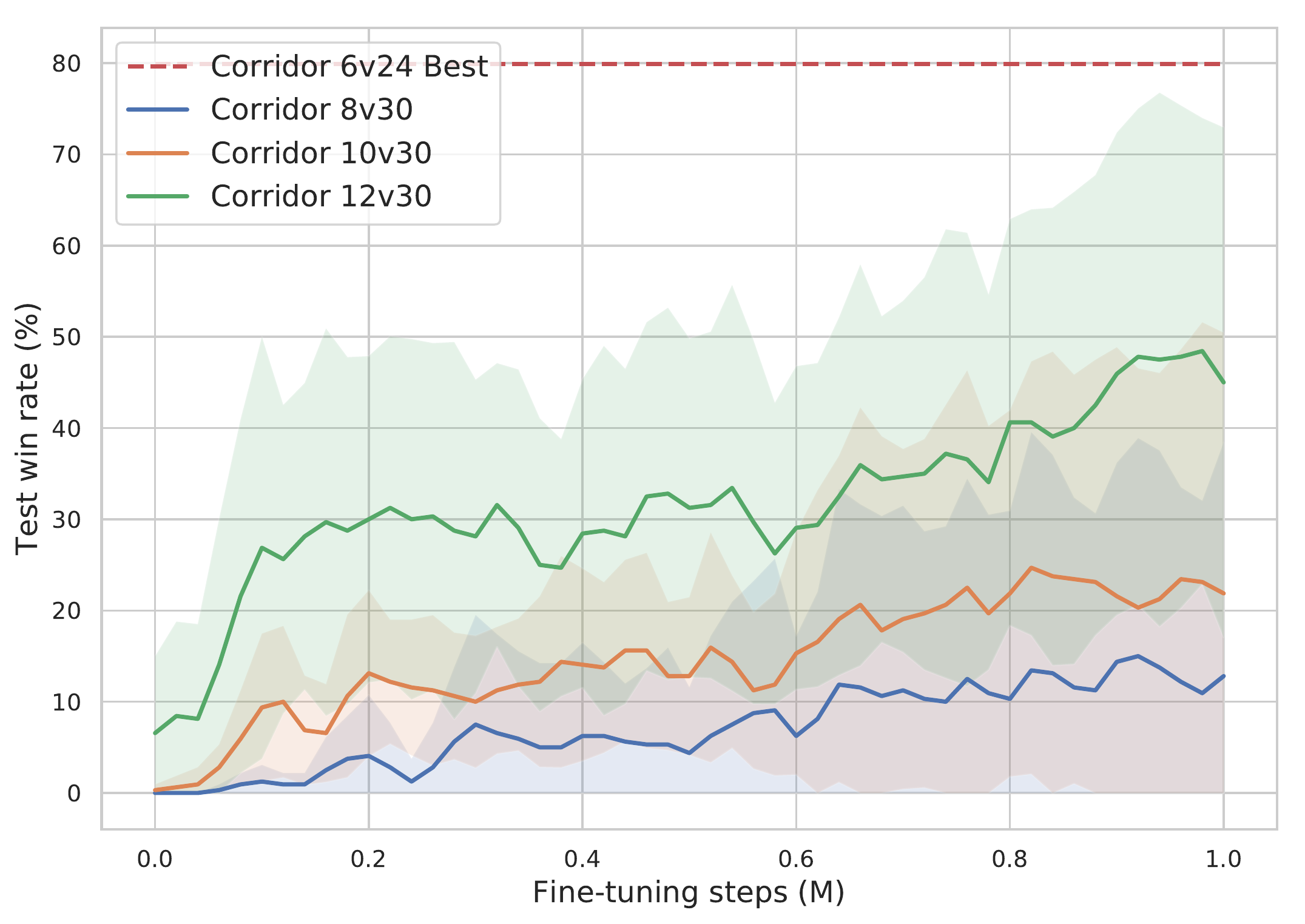}
\caption{Win rates achieved by combined RODE + GraphMIX agents, trained originally on the 6v24 \texttt{corridor} map, and fine-tuned on larger \texttt{corridor} maps with more agents and enemies.}
\label{fig:finetuning_corridor}
\end{figure}
In~\citep{wang2021rode}, it was shown that agents trained using the RODE framework on the original 6v24 \texttt{corridor} map can be used in extended \texttt{corridor} scenarios with more allied/enemy units and achieve an acceptable win rate. Such a ``rapid transfer'' was possible because RODE is insensitive to the dimensionality of the action space due to the specific structure of the role and action agents. Nevertheless, there still existed many scenarios (especially ones in which the enemy units far outnumbered the allied agents), where the transferred policies did not work well. As RODE uses QMIX for role and action value function factorization, fine-tuning the agents is not possible since the input size of the mixing networks grows with the number of agents.

In contrast, as the mixing GNN in GraphMIX is size-invariant, the combined RODE + GraphMIX approach enjoys insensitivity to both the number of agents and the number of actions. This enables us to fine-tune the agents through the mixing GNN on scenarios in which the number of agents is different than that of the training scenario. Figure~\ref{fig:finetuning_corridor} shows the mean win rates achieved by combined RODE + GraphMIX, originally trained on the 6v24 \texttt{corridor} map, upon being fine-tuned on larger \texttt{corridor} maps with 8-12 agents and 30 enemies, where the shaded areas represent the standard deviation. As the figure demonstrates, the fine-tuned agents can improve their win rates when fine-tuned for as few as $10^6$ steps, indicating the scalability of the proposed graph-based value decomposition method due to its size invariance. More details on this experiment can be found in Appendix~\ref{appx:implementation}..

\section{Conclusion}
\label{sec:conclusion}
We introduced GraphMIX, a novel approach to decompose joint state-action value functions in multi-agent deep reinforcement learning (MARL) using a graph neural network formulation under the centralized training and decentralized execution paradigm. Our proposed method allows for a more explicit representation of agent-to-agent relationships by leveraging an attention-based graph topology that models the dynamics between the agents as the episodes progress. To build upon the factorized state-action value function's implicit assignment of global reward, we defined additional per-agent loss terms derived from the output node embeddings of the graph neural network, which explicitly divide the global reward to individual agents. Experiments in the StarCraft Multi-Agent Challenge (SMAC) benchmark demonstrated improved performance over the state-of-the-art multi-agent deep reinforcement learning algorithms across multiple hard and super hard scenarios. We further showed the scalability of GraphMIX, when combined with a hierarchical role-based MARL architecture, by enabling the agents trained on a specific scenario to be fine-tuned on test scenarios with higher numbers of agents.



\vskip 0.2in
\bibliography{MARL_GraphMIX}



\appendix




\section{Implementation Details}\label{appx:implementation}

\subsection{Shared Parameters}
We use a gated recurrent unit (GRU) for each of the decentralized agents in all algorithms with $D=64$ hidden units. Each agent uses an $\epsilon$-greedy policy, where the probability of random actions decays from 100\% to 5\% over $5\times10^{4}$ time steps, except for three super hard maps (\texttt{6h\_vs\_8z}, \texttt{3s5z\_vs\_3s6z}, and \texttt{27m\_vs\_30m}), where the decay happens over $5\times10^{5}$ time steps to allow further exploration~\cite{wang2021rode}. The agents are trained over consecutive episodes, where at the end of each episode, a batch of 32 episodes is randomly sampled from an experience buffer of size 5000 episodes for a round of training. The learning rate is fixed at $5\times10^{-4}$. To stabilize training, double Q-learning is used~\citep{NIPS2010_3964, 10.5555/3016100.3016191}, where the target agent Q-network and the target mixing network parameters are replaced with those of their main counterparts every 200 episodes. Moreover, every $2\times10^4$ time steps, training is paused and the agents are evaluated on a set of 32 test episodes.

\subsection{GraphMIX}
The encoding mechanism $\phi$ for the attention coefficients in~\eqref{eq:attn} is implemented using a single-layer mapping followed by a non-linearity; i.e., $\phi(c_m) = \sigma(Bc_m)$, where $\sigma(\cdot)$ denotes a non-linearity and $B\in\mathbb{R}^{D' \times 64}$ denotes the encoding matrix. We use the exponential linear unit (ELU) as the non-linearity and set $D'=16$.

We consider a single 32-dimensional hidden layer for the mixing GNN and average pooling readout at the output, and for the GIN combining operation in~\eqref{eq:combining_simplified}, we use an MLP with a single hidden layer of size 16 and rectified linear unit (ReLU) non-linearity. The hypernetworks are also MLPs, each with a single hidden layer of size 64 and ReLU non-linearity.

Over the course of an episode, the agents might get killed by the opponent team. In that case, we isolate the dead agents from the other nodes in graph $G$, and we remove them from calculations of the attention weights and the GNN operations entirely. Specifically, the readout operation in~\eqref{eq:readout_GNN}, the effective reward fraction calculation in~\eqref{eq:effective_rewards}, and the softmax operation for deriving the edge weights in~\eqref{eq:attn} are constrained to the agents that are still alive in the corresponding time steps.

\subsection{QMIX}
We use a similar MLP-based architecture for the mixing network in QMIX to its counterpart in GraphMIX; i.e., an MLP with a single hidden layer, 32 neurons per hidden layer, and ELU non-linearity.

\subsection{Weighted QMIX}
We use the Optimistically-Weighted version of the Weighted QMIX algorithm. On most of the maps, the mixing network $Q^*$ in equation (9) of \cite{rashid2020weighted} is implemented using a 3-layer MLP, with each hidden layer having 256 neurons and ReLU non-linearity. For the \texttt{corridor} map, the weight matrix in the first layer of $Q^*$ is derived by applying softmax on the output of a hypernetwork with a single 64-dimensional hidden layer that takes the environment state as input. Moreover, the parameter $\alpha$ in equation (5) of \cite{rashid2020weighted} is set to $0.5$ on most maps, except for \texttt{corridor}, where it is set to $0.75$.

\subsection{RODE}
We mostly use similar hyperparameters to those used in~\cite{wang2021rode} for the number of role clusters and role interval. Specifically, on the maps with heterogeneous enemy units, the number of role clusters is set to $k=5$, while on most maps with homogeneous enemy units, we set the number of role clusters to $k=3$. We also set the role interval to $c=5$.

In our experiments on the \texttt{6h\_vs\_8z} map, we were unable to reproduce similar win rates for RODE to those reported in~\cite{wang2021rode} using $k=3$ role clusters and a role interval of $c=7$. We conducted multiple experiments using the following set of hyperparameters on this map,
\begin{align*}
(k, c)\in\{(3,3), (3,5), (3,6), (3,7), (3,8), (3,10), (5,3), (5,5), (5,7), (7,7)\},
\end{align*}
and found that $(k, c)=(5,5)$ led to the best results, which are still below the win rates reported in~\cite{wang2021rode}. We, therefore, set $k=c=5$ for our experiments on this map.

\subsection{RODE + GraphMIX (Main Experiments)}
As mentioned in Section~\ref{sec:baselines}, the combined RODE + GraphMIX approach introduces a new role local loss hyperparameter $\lambda_{\mathsf{role}, \mathsf{local}}$ aside from the original local loss hyperparameter $\lambda_{\mathsf{local}}$ in~\eqref{eq:aggregate_loss}. Table~\ref{tab:local_loss_coeffs} shows the local loss hyperparameters we use for the combined RODE + GraphMIX method on the nine maps.

\aboverulesep=0ex
\belowrulesep=0ex
\begin{table}[h!]
\centering
\noindent\makebox[\textwidth]{
\rowcolors{1}{Gray}{}
\setlength\tabcolsep{10pt}
\begin{tabular}{l|cc}
\cmidrule[1.5pt]{1-3}
Map & $\lambda_{\mathsf{local}}$ & $\lambda_{\mathsf{role}, \mathsf{local}}$ \\ 
\cline{1-3}
\texttt{6h\_vs\_8z} & 1 & 1 \\
\texttt{corridor} & 0 & 1 \\
\texttt{MMM2} & 0 & 0 \\
\texttt{3s5z\_vs\_3s6z} & 0 & 1 \\
\texttt{27m\_vs\_30m} & 0 & 0 \\
\texttt{2c\_vs\_64zg} & 0 & 0 \\
\texttt{5m\_vs\_6m} & 0 & 0 \\
\texttt{bane\_vs\_bane} & 0 & 0 \\
\texttt{3s\_vs\_5z} & 1 & 0 \\
\cmidrule[1.5pt]{1-3}
\end{tabular}}
\caption{Local loss coefficients used for the combined RODE + GraphMIX approach.}
\label{tab:local_loss_coeffs}
\end{table}

Due to the different agent structures in RODE, we use the projected hidden states, i.e., $z_{\tau_i}$ in equations (3) and (5) of~\cite{wang2021rode} instead of the hidden states $c_m$ to compute the attention coefficients in~\eqref{eq:attn} for both the role and action mixing GNNs.

\subsection{RODE + GraphMIX (Fine-Tuning Experiments on \texttt{corridor})}
We take a similar approach to that of the policy transfer experiments in~\cite{wang2021rode}, where on each of the new maps, we first train an action encoder for the new action space over $5\times10^4$ steps. Then, for each new action that was not present in the original 6v24 map, we find the 5 nearest neighbors among the old actions in the action embedding space. We use the average of those neighboring action embeddings as the embedding of the new action, and add that action to the union of roles (i.e., clusters) to which those neighboring actions belong. Note  that we do not update the role representations and old action representations based on the new action embeddings.

Moreover, similarly to~\cite{wang2021rode}, to keep the per-agent observation length unchanged across the original and extended \texttt{corridor} maps, we only let each agent observe the closest $5$ allied units and $24$ enemy units (if they are within its visible range). Moreover, to keep the state space dimensionality constant across all the maps, we include the information about the $6$ allied units with the smallest mean distances to the enemy units, and $24$ enemy units with the smallest mean distances to the allied units. Keeping the state vector length fixed is necessary as we still need to use the global environment state to fine-tune the mixing GNNs on the new maps.

\subsection{Hardware}
We run most of our experiments on a Linux machine with a $2.40$ GHz Intel\textsuperscript{\tiny \textregistered} Xeon\textsuperscript{\tiny \textregistered} E5-2680 v4 CPU and two $16$ GB NVIDIA\textsuperscript{\tiny \textregistered} Tesla\textsuperscript{\tiny \textregistered} P100 GPUs, while for the Weighted QMIX experiments on two of the maps, namely \texttt{27m\_vs\_30m} and \texttt{bane\_vs\_bane}, due to higher memory requirements, we use a more powerful Linux machine with a $2.70$ GHz Intel\textsuperscript{\tiny \textregistered} Xeon\textsuperscript{\tiny \textregistered} Platinum 8280 CPU and four $48$ GB NVIDIA\textsuperscript{\tiny \textregistered} Quadro RTX 8000 GPUs.

\section{Ablation Results on The Impact of The Local Loss Terms in The Aggregate Loss}\label{appx:loss_ablation}
\begin{figure*}[t]
\centering
\includegraphics[width=\textwidth]{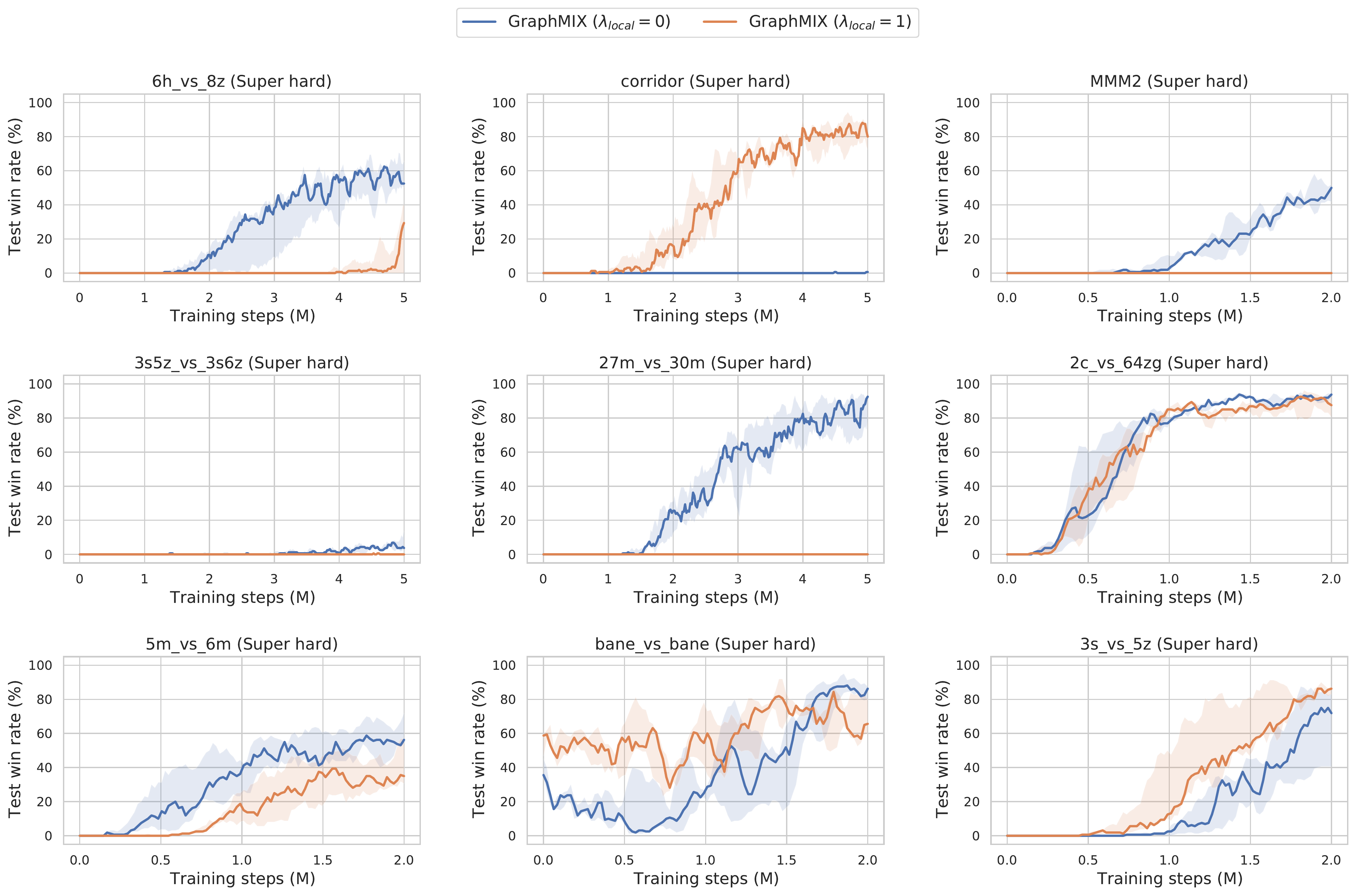}
\caption{Comparison of the performance of GraphMIX on all the nine hard and super hard maps with and without the local loss terms in~\eqref{eq:aggregate_loss}.}
\label{fig:ablation_localloss_allmaps}
\end{figure*}

Figure~\ref{fig:ablation_localloss_allmaps} shows the impact of including or excluding the local loss terms in~\eqref{eq:aggregate_loss} (i.e., $\lambda_{\mathsf{local}}\in\{1,0\}$) on the performance of GraphMIX on all the nine hard and super hard maps. As the figure shows, while including the local per-agent losses is essential on some of the maps, e.g., \texttt{corridor}, the performance of GraphMIX improves using only the global loss on the rest of the maps. Therefore, it is important that this hyperparameter be tuned on any environment of interest so as to maximize the performance gains of GraphMIX.

\end{document}